\documentclass[11pt,a4paper]{article}

\usepackage{times}
\usepackage{latexsym}
\usepackage[hide links]{hyperref}
\usepackage{booktabs}
\usepackage{amsmath}
\usepackage{xspace}
\usepackage{booktabs}
\usepackage{makecell}
\usepackage{mathtools}
\usepackage{dsfont}
\usepackage{stix}
\usepackage{comment}
\usepackage{naaclhlt2018}

\aclfinalcopy % Uncomment this line for the final submission
\title{Explainable Prediction of Medical Codes from Clinical Text}
   \author{James Mullenbach, Sarah Wiegreffe, Jon Duke, Jimeng Sun, Jacob Eisenstein \\
     Georgia Institute of Technology \\
     \texttt{\{jmullenbach3, swiegreffe6, jon.duke\}@gatech.edu} \\ \texttt{jsun@cc.gatech.edu, jacobe@gatech.edu} \\
   }
\date{}

\renewcommand{\vec}[1]{\boldsymbol{#1}}
\newcommand{\mat}[1]{\boldsymbol{#1}}
\newcommand{\reals}{\mathbb{R}}
\newcommand{\demb}{d_{e}}
\newcommand{\dconv}{d_c}
\newcommand{\pate}{precision$@8$}
\newcommand{\numlabels}{\ensuremath |\mathcal{L}|}
\newcommand{\sumlabels}{\sum_{\ell = 1}^{\numlabels}}

\DeclareMathOperator*{\argmax}{arg\,max}

\begin{document}
\maketitle

\begin{abstract}
Clinical notes are text documents that are created by clinicians for each patient encounter. They are typically accompanied by medical codes, which describe the diagnosis and treatment. Annotating these codes is labor intensive and error prone; furthermore, the connection between the codes and the text is not annotated, obscuring the reasons and details behind specific diagnoses and treatments. We present an attentional convolutional network that predicts medical codes from clinical text. Our method aggregates information across the document using a convolutional neural network, and uses an attention mechanism to select the most relevant segments for each of the thousands of possible codes. The method is accurate, achieving \pate{} of 0.71 and a Micro-F1 of 0.54, which are both better than the prior state of the art. Furthermore, through an interpretability evaluation by a physician, we show that the attention mechanism identifies meaningful explanations for each code assignment.
\end{abstract}

\section{Introduction}
Clinical notes are free text narratives generated by clinicians during patient encounters.
They are typically accompanied by a set of metadata codes from the International Classification of Diseases (ICD), which present a standardized way of indicating diagnoses and procedures that were performed during the encounter. ICD codes have a variety of uses, ranging from billing to predictive modeling of patient state~\cite{choi2016doctor, ranganath2015survival,denny2010phewas,avati2017improving}. Because manual coding is time-consuming and error-prone, automatic coding has been studied since at least the 1990s~\cite{de1998hierarchical}. The task is difficult for two main reasons. First, the label space is very high-dimensional, with over 15,000 codes in the ICD-9 taxonomy, and over 140,000 codes combined in the newer ICD-10-CM and ICD-10-PCS taxonomies \cite{ICD10}. Second, clinical text includes irrelevant information, misspellings and non-standard abbreviations, and a large medical vocabulary. These features combine to make the prediction of ICD codes from clinical notes an especially difficult task, for computers and human coders alike~\cite{birman2005accuracy}.

\begin{table*}
  \centering
  \small
\begin{tabular}{lp{4.8in}}

  \multicolumn{2}{l}{\textbf{934.1}: ``Foreign body in main bronchus''} \\
  \toprule
CAML (HI) & \textit{...line placed bronchoscopy performed showing} \textbf{large mucus plug on} \textit{the left on transfer to...} \\
Cosine Sim & \textit{...also needed medication to help} \textbf{your body maintain your} \textit{blood pressure after receiving iv...} \\
CNN & \textit{...found to have a large} \textbf{lll lingular pneumonia on} \textit{chest x ray he was...} \\
Logistic Regression & \textit{...impression confluent consolidation involving nearly} \textbf{the entire left lung} \textit{with either bronchocentric or vascular...} \\[1em]

\multicolumn{2}{l}{\textbf{442.84}: ``Aneurysm of other visceral artery''} \\
  \toprule
CAML (I) & \textit{...and gelfoam embolization of right} \textbf{hepatic artery branch pseudoaneurysm} \textit{coil embolization of the gastroduodenal...} \\
Cosine Sim & \textit{...coil embolization of the gastroduodenal} \textbf{artery history of present} \textit{illness the pt is a...} \\
CNN & \textit{...foley for hemodynamic monitoring and} \textbf{serial hematocrits angio was} \textit{performed and his gda was...} \\
Logistic Regression (I) & \textit{...and gelfoam embolization of right} \textbf{hepatic artery branch pseudoaneurysm} \textit{coil embolization of the gastroduodenal...} \\[1em]

\multicolumn{2}{l}{\textbf{428.20}: ``Systolic heart failure, unspecified''} \\ \toprule
CAML & \textit{...no mitral valve prolapse moderate} \textbf{to severe mitral regurgitation} \textit{is seen the tricuspid valve...} \\
Cosine Sim & \textit{...is seen the estimated pulmonary} \textbf{artery systolic pressure is} \textit{normal there is no pericardial...} \\
CNN & \textit{...and suggested starting hydralazine imdur} \textbf{continue aspirin arg admitted} \textit{at baseline cr appears patient...} \\
Logistic Regression (HI) & \textit{...anticoagulation monitored on tele pump} \textbf{systolic dysfunction with ef} \textit{of seen on recent echo...}

\end{tabular}
\caption{Presentation of example qualitative evaluations. In real evaluation, system names generating the 4-gram are not given. An `I' marking indicates a snippet evaluated as informative, and `HI' indicates that it is highly informative; see \autoref{sec:qual} for more details. %One good example of attn, one with multiple good explanations, one with bad attention, but was still attending to relevant things
}
\label{tab:qual_example}
\end{table*}

In this application paper, we develop convolutional neural network (CNN)-based methods for automatic ICD code assignment based on text discharge summaries from intensive care unit (ICU) stays. To better adapt to the multi-label setting, we employ a per-label attention mechanism, which allows our model to learn distinct document representations for each label. We call our method \textbf{C}onvolutional \textbf{A}ttention for \textbf{M}ulti-\textbf{L}abel classification (CAML). Our model design is motivated by the conjecture that important information correlated with a code's presence may be contained in short snippets of text which could be anywhere in the document, and that these snippets likely differ for different labels. To cope with the large label space, we exploit the textual descriptions of each code to guide our model towards appropriate parameters: in the absence of many labeled examples for a given code, its parameters should be similar to those of codes with similar textual descriptions. 

We evaluate our approach on two versions of MIMIC~\cite{johnson2016mimic}, an open dataset of ICU medical records. Each record includes a variety of narrative notes describing a patient's stay, including diagnoses and procedures.
Our approach substantially outperforms previous results on medical code prediction on both MIMIC-II and MIMIC-III datasets. 

We consider applications of this work in a decision support setting. Interpretability is important for any decision support system, especially in the medical domain. The system should be able to explain why it predicted each code; even if the codes are manually annotated, it is desirable to explain what parts of the text are most relevant to each code. These considerations further motivate our per-label attention mechanism, which assigns importance values to $n$-grams in the input document, and which can therefore provide explanations for each code, in the form of extracted snippets of text from the input document. We perform a human evaluation of the quality of the explanations provided by the attention mechanism, asking a physician to rate the informativeness of a set of automatically generated explanations.\footnote{Our code, data splits, and pre-trained models are available at \url{github.com/jamesmullenbach/caml-mimic}.}

\section{Method}

We treat ICD-9 code prediction as a multilabel text classification problem~\cite{mccallum1999multi}.\footnote{We focus on codes from the ICD-9 taxonomy, rather than the more recent ICD-10, for the simple reason that this is the version of ICD used in the MIMIC datasets.} Let $\mathcal{L}$ represent the set of ICD-9 codes; the labeling problem for instance $i$ is to determine $y_{i,\ell} \in \{0,1\}$ for all $\ell \in \mathcal{L}$. We train a neural network which passes text through a convolutional layer to compute a base representation of the text of each document~\cite{kim2014convolutional}, and makes $|\mathcal{L}|$ binary classification decisions. Rather than aggregating across this representation with a pooling operation, we apply an attention mechanism to select the parts of the document that are most relevant for each possible code. These attention weights are then applied to the base representation, and the result is passed through an output layer, using a sigmoid transformation to compute the likelihood of each code. 
We employ a regularizer to encourage each code's parameters to be similar to those of codes with similar textual descriptions. We now describe each of these elements in more detail.

\subsection{Convolutional architecture}
At the base layer of the model, we have {$\demb$-dimensional} pre-trained embeddings for each word in the document, which are horizontally concatenated into the matrix $\mat{X} = [\vec{x}_1, \vec{x}_2, \ldots, \vec{x}_N]$, where $N$ is the length of the document. 
Adjacent word embeddings are combined using a convolutional filter $\mat{W}_c \in \reals^{k \times \demb \times \dconv}$, where $k$ is the filter width, $\demb$ the size of the input embedding, and $\dconv$ the size of the filter output. At each step $n$, we compute
\begin{equation}
  \vec{h}_n = g(\mat{W}_c \ast \vec{x}_{n:n+k-1} + \vec{b}_c),
\end{equation}
where $\ast$ denotes the convolution operator, $g$ is an element-wise nonlinear transformation, and $\vec{b}_c \in \mathbb{R}^{\dconv}$ is the bias. 
We additionally pad each side of the input with zeros so that the resulting matrix $\mat{H}$ has dimension $\reals^{\dconv \times N}$.

\subsection{Attention}\label{sec:attn}
After convolution, the document is represented by the matrix $\mat{H} \in \reals^{\dconv \times N}$. It is typical to reduce this matrix to a vector by applying pooling across the length of document, by selecting the maximum or average value at each row~\citep{kim2014convolutional}. However, our goal is to assign multiple labels (i.e., medical codes) for each document, and different parts of the base representation may be relevant for different labels. For this reason, we apply a per-label attention mechanism. An additional benefit is that it selects the $k$-grams from the text that are most relevant to each predicted label.

Formally, for each label $\ell$, we compute the matrix-vector product, $\mat{H}^{\top} \vec{u}_\ell$, where $\vec{u}_\ell \in \reals^{\dconv}$ is a vector parameter for label $\ell$. 
We then pass the resulting vector through a softmax operator, obtaining a distribution over locations in the document,
\begin{equation}
  \vec{\alpha}_\ell = \text{SoftMax}(\mat{H}^{\top} \vec{u}_\ell),
\end{equation}
where $\text{SoftMax}(\vec{x}) = \frac{\exp (\vec{x})}{\sum_i \exp (x_i)}$, and $\exp(\vec{x})$ is the element-wise exponentiation of the vector $\vec{x}$. The attention vector $\vec{\alpha}$ is then used to compute vector representations for each label,
\begin{equation}
  \vec{v}_\ell = \sum^N_{n=1} \alpha_{\ell, n} \vec{h}_n.
\end{equation}

As a baseline model, we instead use max-pooling to compute a single vector $\vec{v}$ for all labels,
\begin{equation}
  v_j = \max_{n} h_{n,j}.
\label{eq:maxpool}
\end{equation}

\subsection{Classification}
\begin{figure}
\centering
\includegraphics[width=0.5\textwidth]{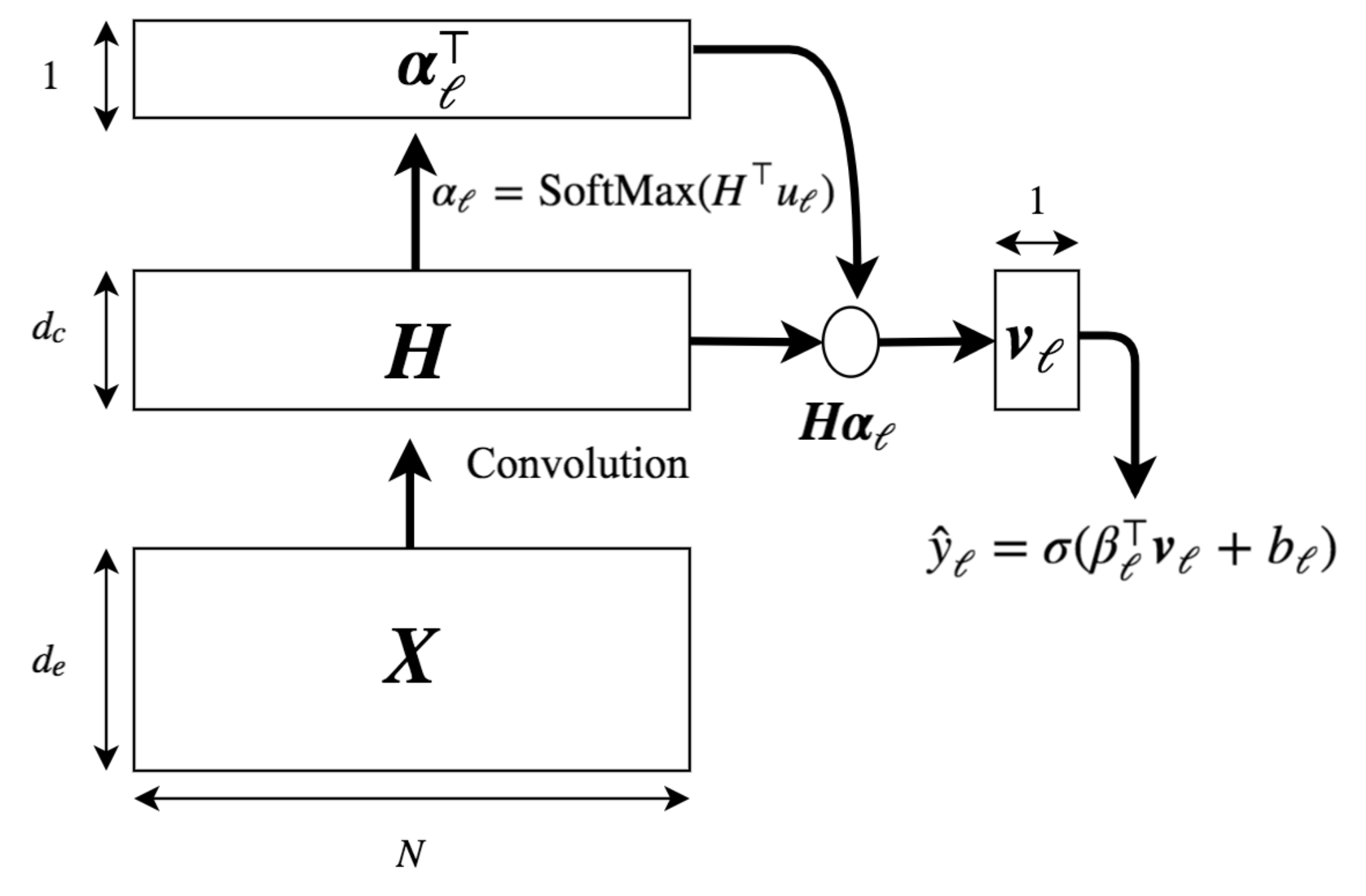}
\caption{CAML architecture with per-label attention shown for one label. In a max-pooling architecture, $\vec{H}$ is mapped directly to the vector $\vec{v}_\ell$ by maximizing over each dimension.}
\label{fig:conv_attn}
\end{figure}

Given the vector document representation $\vec{v}_\ell$, we compute a probability for label $\ell$ using another linear layer and a sigmoid transformation:
\begin{equation}
  \hat{y}_\ell = \sigma(\vec{\beta}_\ell^\top \vec{v}_\ell + b_{\ell}),
  \label{eq:prediction}  
\end{equation}
where $\vec{\beta}_\ell \in \reals^{\dconv}$ is a vector of prediction weights, and $b_\ell$ is a scalar offset. The overall model is illustrated in \autoref{fig:conv_attn}. 

\subsection{Training}
The training procedure minimizes the binary cross-entropy loss,
\begin{align}
\begin{split}
  L_{\textrm{BCE}}(\mat{X}, \vec{y}) =  -\sum_{\ell=1}^{\mathcal{L}} &{} y_\ell\log(\hat{y}_\ell)\\
  &{} + (1-y_\ell)\log(1-\hat{y}_\ell),
\end{split}
    \label{eq:bce}
\end{align}
plus the L2 norm of the model weights, using the Adam optimizer~\cite{kingma2014adam}.

\subsection{Embedding label descriptions}\label{sec:desc_embed}

Due to the dimensionality of the label space, many codes are rarely observed in the labeled data.
To improve performance on these codes, we use text descriptions of each code from the \newcite{ICD10}.
Examples can be found in \autoref{tab:qual_example}, next to the code numbers.
We use these descriptions to build a secondary module in our network that learns to embed 
them as vectors. These vectors are then used as the target of regularization on the model parameters $\vec{\beta}_{\ell}$. If code $\ell$ is rarely observed in the training data, this regularizer will encourage its parameters to be similar to those of other codes with similar descriptions.

The code embedding module consists of a max-pooling CNN architecture. Let $\vec{z}_\ell$ be a max-pooled vector, obtained by passing the description for code $\ell$ into the module. Let $n_y$ be the number of true labels in a training example. We add the following regularizing objective to our loss $L$,
\begin{align}
L(\mat{X}, \vec{y}) = L_{\textrm{BCE}} + \lambda\frac{1}{n_y}\sum_{\ell:y_\ell = 1}^{\mathcal{L}}\|\vec{z}_\ell - \vec{\beta}_\ell\|_2,
\label{eq:desc_reg}
\end{align}
where $\lambda$ is a tradeoff hyperparameter that calibrates the performance of the two objectives. We call this model variant Description Regularized-CAML (DR-CAML).

\section{Evaluation of code prediction}
This section evaluates the accuracy of code prediction, comparing our models against several competitive baselines.

\subsection{Datasets}
\begin{table*}
\centering
\begin{tabular}{llll}
\toprule
& \textbf{MIMIC-III full} & \textbf{MIMIC-III 50} & \textbf{MIMIC-II full} \\ \midrule
\# training documents                &  47,724               &    8,067                & 20,533                 \\
Vocabulary size             & 51,917                & 51,917                  & 30,688                 \\
Mean \# tokens per document & 1,485                 & 1,530                  & 1,138                  \\
Mean \# labels per document & 15.9                 & 5.7                   & 9.2                    \\
Total \# labels             & 8,922                   & 50                  & 5,031 \\ \bottomrule
\end{tabular}
\caption{Descriptive statistics for MIMIC discharge summary training sets.}
\label{tab:stats}
\end{table*}

MIMIC-III~\cite{johnson2016mimic} is an open-access dataset of text and structured records from a hospital ICU. Following previous work, we focus on discharge summaries, which condense information about a stay into a single document. In MIMIC-III, some admissions have addenda to their summary, which we concatenate to form one document.

Each admission is tagged by human coders with a set of ICD-9 codes, describing both diagnoses and procedures which occurred during the patient's stay. There are 8,921 unique ICD-9 codes present in our datasets, including 6,918 diagnosis codes and 2,003 procedure codes. Some patients have multiple admissions and therefore multiple discharge summaries; we split the data by patient ID, so that no patient appears in both the training and test sets.

In this full-label setting, we use a set of 47,724 discharge summaries from 36,998 patients for training, with 1,632 summaries and 3,372 summaries for validation and testing, respectively. 

\paragraph{Secondary evaluations}
For comparison with prior work, we also follow \newcite{shi2017towards} and train and evaluate on a label set consisting of the 50 most frequent labels. In this setting, we filter each dataset down to the instances that have at least one of the top 50 most frequent codes, and subset the training data to equal the size of the training set of \newcite{shi2017towards}, resulting in 8,067 summaries for training, 1,574 for validation, and 1,730 for testing.

We also run experiments with the MIMIC-II dataset, to compare with prior work by \newcite{baumel2017multi} and \newcite{perotte2013diagnosis}. We use the train/test split of \newcite{perotte2013diagnosis}, which consists of 20,533 training examples and 2,282 testing examples. Detailed  statistics for the three settings are summarized in \autoref{tab:stats}.

\paragraph{Preprocessing}
We remove tokens that contain no alphabetic characters (e.g., removing ``500'' but keeping ``250mg''), lowercase all tokens, and replace tokens that appear in fewer than three training documents with an `UNK' token. We pretrain word embeddings of size $d_e = 100$ using the word2vec CBOW method~\cite{mikolov2013efficient} on the preprocessed text from all discharge summaries. All documents are truncated to a maximum length of 2500 tokens.

\subsection{Systems}
We compare against the following baselines:
\begin{itemize}
\item a single-layer one-dimensional convolutional neural network \cite{kim2014convolutional};
\item a bag-of-words logistic regression model;
\item a bidirectional gated recurrent unit (Bi-GRU).\footnote{Our pilot experiments found that GRU was stronger than long short-term memory (LSTM) for this task.}
\end{itemize}
For the CNN and Bi-GRU, we initialize the embedding weights using the same pretrained word2vec vectors that we use for the CAML models. All neural models are implemented using PyTorch\footnote{\url{https://github.com/pytorch/pytorch}}. The logistic regression model consists of $|\mathcal{L}|$ binary one-vs-rest classifiers acting on unigram bag-of-words features for all labels present in the training data. If a label is not present in the training data, the model will never predict it in the held-out data. 

\paragraph{Parameter tuning}
We tune the hyperparameters of the CAML model and the neural baselines using the Spearmint Bayesian optimization package~\citep{NIPS2012_4522, NIPS2013_5086}.\footnote{\url{https://github.com/HIPS/Spearmint}}
We allow Spearmint to sample parameter values for the L2 penalty on the model weights $\rho$ and learning rate $\eta$, as well as filter size $k$, number of filters $d_c$, and dropout probability $q$ for the convolutional models, and number of hidden layers $s$ of dimension $v$ for the Bi-GRU, using \pate{} on the MIMIC-III full-label validation set as the performance measure. We use these parameters for DR-CAML as well, and port the optimized parameters to the MIMIC-II full-label and MIMIC-III 50-label models, and manually fine-tune the learning rate in these settings. We select $\lambda$ for DR-CAML based on pilot experiments on the validation sets. Hyperparameter tuning is summarized in \autoref{tab:hyperparams}.
Convolutional models are trained with dropout after the embedding layer. We use a fixed batch size of 16 for all models and datasets. Models are trained with early stopping on the validation set; training terminates after the precision@8 does not improve for 10 epochs, and the model at the time of the highest precision@8 is used on the test set. 

\begin{table}
\centering
  \small
\begin{tabular}{lp{2.4cm}llll}
  \toprule
 & Range & CAML & CNN & Bi-GRU \\ \midrule
$d_c$ & 50-500 & 50 & 500 & -- \\
$k$ & 2-10 & 10 & 4 & -- \\
$q$ & 0.2-0.8 & 0.2 & 0.2 & -- \\
$\rho$ & 0, 0.001, 0.01, 0.1 & 0 & 0 & 0 \\
$\eta$ & 0.0001, 0.0003, 0.001, 0.003 & 0.0001 & 0.003 & 0.003 \\
$s$ & 1-4 & -- & -- & 1 \\
$v$ & 32-512 & -- & -- & 512 \\ \bottomrule
\end{tabular}
\caption{Hyperparameter ranges and optimal values for each neural model selected by Spearmint.
  }
\label{tab:hyperparams}
\end{table}

\subsection{Evaluation Metrics}
To facilitate comparison with both future and prior work, we report a variety of metrics, focusing on the micro-averaged and macro-averaged F1 and area under the ROC curve (AUC). Micro-averaged values are calculated by treating each (text, code) pair as a separate prediction. Macro-averaged values, while less frequently reported in the multi-label classification literature, are calculated by averaging metrics computed per-label. For recall, the metrics are distinguished as follows:
\begin{align}
\text{Micro-R} = &{} \frac{\sumlabels \text{TP}_{\ell}}{\sumlabels \text{TP}_{\ell} + \text{FN}_{\ell}}\\
  \text{Macro-R} = &{} \frac{1}{\numlabels} \sumlabels \frac{\text{TP}_{\ell}}{\text{TP}_{\ell} + \text{FN}_{\ell}},
\end{align}
where TP denotes true positive examples and FN denotes false negative examples. Precision is computed analogously. The macro-averaged metrics place much more emphasis on rare label prediction. 

We also report precision at $n$ (denoted as `P@n'), which is the fraction of the $n$ highest-scored labels that 
are present in the ground truth. This is motivated by the potential use case as a decision support application, in which a user is presented with a fixed number of predicted codes to review. In such a case, it is more suitable to select a model with high precision than high recall. We choose $n = 5$ and $n = 8$ to compare with prior work~\cite{vani2017grounded,prakash2017condensed}. For the MIMIC-III full label setting, we also compute precision@15, which roughly corresponds to the average number of codes in MIMIC-III discharge summaries (\autoref{tab:stats}). 

\begin{table*}
\centering
\begin{tabular}{lll|llll|ll}
\toprule
& \multicolumn{2}{c}{AUC} & \multicolumn{4}{c}{F1} & \multicolumn{2}{c}{P@n}  \\
Model & Macro & Micro & Macro & Micro & Diag & Proc & 8 & 15 \\ \midrule
Scheurwegs et. al \shortcite{scheurwegs2017selecting} & -- & -- & -- & -- & 0.428 & 0.555 & -- & --  \\[1em]
Logistic Regression & 0.561 & 0.937 & 0.011 & 0.272 & 0.242 & 0.398 & 0.542 & 0.411 \\
CNN & 0.806 & 0.969 & 0.042 & 0.419 & 0.402 & 0.491 & 0.581 & 0.443 \\
Bi-GRU & 0.822 & 0.971 & 0.038 & 0.417 & 0.393 & 0.514 & 0.585 & 0.445 \\ \midrule
CAML & 0.895 & \textbf{0.986}* & \textbf{0.088} & \textbf{0.539}* & \textbf{0.524}* & \textbf{0.609}* & \textbf{0.709}* & \textbf{0.561}* \\
DR-CAML & \textbf{0.897} & 0.985 & 0.086 & 0.529 & 0.515 & 0.595 & 0.690 & 0.548 \\
\bottomrule
\end{tabular}
\caption{Results on MIMIC-III full, 8922 labels. Here, ``Diag'' denotes Micro-F1 performance on diagnosis codes only, and ``Proc'' denotes Micro-F1 performance on procedure codes only. Here and in all tables, (*) by the bold (best) result indicates significantly improved results compared to the next best result, $p < 0.001$.}
\label{tab:mimic3-full}
\end{table*}

\subsection{Results}

Our main quantitative evaluation involves predicting the full set of ICD-9 codes based on the text of the MIMIC-III discharge summaries. These results are shown in \autoref{tab:mimic3-full}. The CAML model gives the strongest results on all metrics. Attention yields substantial improvements over the ``vanilla'' convolutional neural network (CNN). The recurrent Bi-GRU architecture is comparable to the vanilla CNN, and the logistic regression baseline is substantially worse than all neural architectures. The best-performing CNN model has 9.86M tunable parameters, compared with 6.14M tunable parameters for CAML. This is due to the hyperparameter search preferring a larger number of filters for the CNN. 
Finally, we observe that the DR-CAML performs worse on most metrics than CAML, with a tuned regularization coefficient of $\lambda = 0.01$. 

Among prior work, only \newcite{scheurwegs2017selecting} evaluate on the full ICD-9 code set for MIMIC-III. Their reported results distinguished between diagnosis codes and procedure codes. The CAML models are stronger on both sets. Additionally, our method does not make use of any external information or structured data, while Scheurwegs et al. use structured data and various medical ontologies in their text representation.

We feel that precision@8 is the most informative of the metrics, as it measures the ability of the system to return a small high-confidence subset of codes. Even with a space of thousands of labels, our models achieve relatively high precision: of the eight most confident predictions, on average 5.5 are correct. It is also apparent how difficult it is to achieve high Macro-F1 scores, due to the metric's emphasis on rare-label performance. To put these results in context, a hypothetical system that performs perfectly on the 500 most common labels, and ignores all others, would achieve a Macro-F1 of 0.052 and a Micro-F1 of 0.842. 

\paragraph{Secondary evaluations} 
To compare with prior published work, we also evaluate on the 50 most common codes in MIMIC-III (\autoref{tab:mimic3-50}), and on MIMIC-II (\autoref{tab:mimic2}). We report DR-CAML results on the 50-label setting of MIMIC-III with $\lambda = 10$, and on MIMIC-II with $\lambda = 0.1$, which were determined by grid search on a validation set. The other hyperparameters were left at the settings for the main MIMIC-III evaluation, as described in \autoref{tab:hyperparams}. In the 50-label setting of MIMIC-III, we see strong improvement over prior work in all reported metrics, as well as against the baselines, with the exception of precision@5, on which the CNN baseline performs best. We hypothesize that this is because the relatively large value of $k=10$ for CAML leads to a larger network that is more suited to larger datasets; tuning CAML's hyperparameters on this dataset would be expected to improve performance on all metrics.
\newcite{baumel2017multi} additionally report a micro-F1 score of 0.407 by training on MIMIC-III, and evaluating on MIMIC-II. Our model achieves better performance using only the (smaller) MIMIC-II training set, leaving this alternative training protocol for future work.

\begin{table*}[t]
\centering
\begin{tabular}{lll|ll|l}
\toprule
   & \multicolumn{2}{c}{AUC} & \multicolumn{2}{c}{F1} \\
  Model & Macro & Micro & Macro & Micro & P@5  \\
  \midrule
C-MemNN \cite{prakash2017condensed} & 0.833 & -- & -- & -- & 0.42 \\
\newcite{shi2017towards} & -- & 0.900 & -- & 0.532 & --  \\[1em]
Logistic Regression & 0.829 &            0.864  &         0.477 & 0.533 & 0.546  \\
CNN                 & 0.876 &            0.907  & \textbf{0.576}* &             0.625  & \textbf{0.620} \\
Bi-GRU              & 0.828 & 0.868 & 0.484 & 0.549 & 0.591  \\ \midrule
CAML                & 0.875 & 0.909 & 0.532 & 0.614 & 0.609 \\
DR-CAML      & \textbf{0.884}* & \textbf{0.916} & \textbf{0.576}* & \textbf{0.633} &         0.618 \\
  \bottomrule
\end{tabular}
\caption{Results on MIMIC-III, 50 labels.}
\label{tab:mimic3-50}
\end{table*}

\section{Evaluation of Interpretability}
\label{sec:qual}

We now evaluate the explanations generated by CAML's attention mechanism, in comparison with three alternative heuristics. A physician was presented with explanations from four methods, using a random sample of 100 predicted codes from the MIMIC-III full-label test set. The most important $k$-gram from each method was extracted, along with a window of five words on either side for context. We select $k=4$ in this setting to emulate a span of attention over words likely to be given by a human reader. Examples can be found in \autoref{tab:qual_example}. Observe that the snippets may overlap in multiple words. We prompted the evaluator to select all text snippets which he felt adequately explained the presence of a given code, provided the code and its description, with the option to distinguish snippets as ``highly informative'' should they be found particularly informative over others. 

\subsection{Extracting informative text snippets}

\paragraph{CAML}
The attention mechanism allows us to extract $k$-grams from the text that are most influential in the prediction of each label, by taking the argmax of the SoftMax output $\vec{\alpha}_\ell$.

\paragraph{Max-pooling CNN}
We select the $k$-grams that provide the maximum value selected by max-pooling at least once and weighting by the final layer weights. Defining an argmax vector $\vec{a}$ which results from the max-pooling step as
\begin{align}
\vec{a}_i = \argmax_{j \in \{1,\ldots,m-k+1\}}(\mat{H}_{ij}),
\end{align}
we can compute the importance of position $i$ for label $\ell$,
\begin{align}
\alpha_{i\ell} = \sum_{j:\vec{a}_j=i}^{d_{c}}\beta_{\ell,j} .
\end{align}
We then select the most important $k$-gram for a given label as $\argmax_i\alpha_{i\ell}$.

\paragraph{Logistic regression}
The informativeness of each $k$-gram with respect to label $\ell$ is scored by the sum of the coefficients of the weight matrix for $\ell$, over the words in the $k$-gram. The top-scoring $k$-gram is then returned as the explanation.

\paragraph{Code descriptions} Finally, we calculate a word similarity metric between each stemmed $k$-gram and the stemmed ICD-9 code description. We compute the idf-weighted cosine similarity, with idf weights calculated on the corpus consisting of all notes and relevant code descriptions. We then select the argmax over $k$-grams in the document, breaking ties by selecting the first occurrence. We remove those note-label pairs for which no $k$-gram has a score greater than 0, which gives an ``unfair'' advantage to this baseline.

\subsection{Results}
The results of the interpretability evaluation are presented in \autoref{tab:qual_results}. Our model selects the greatest number of ``highly informative'' explanations, and selects more ``informative'' explanations than both the CNN baseline and the logistic regression model. While the cosine similarity metric also performs well, the examples in \autoref{tab:qual_example} demonstrate the strengths of CAML in extracting text snippets in line with more intuitive explanations for the presence of a code. As noted above, there exist some cases, which we exclude, where the cosine similarity method is unable to provide any explanation, because no $k$-grams in a note have a non-zero similarity for a given label description. This occurs for about 12$\%$ of all note-label pairs in the test set. 
\begin{table*}
\centering
\begin{tabular}{lll|ll|l}
\toprule
 & \multicolumn{2}{c}{AUC} & \multicolumn{2}{c}{F1} &   \\
Model & Macro & Micro & Macro & Micro & P@8   \\
\midrule
Flat SVM \cite{perotte2013diagnosis} & -- & -- & -- & 0.293 & -- \\
HA-GRU \cite{baumel2017multi} & -- & -- & -- & 0.366 & --   \\[1em]
Logistic Regression & 0.690 & 0.934 & 0.025 & 0.314 & 0.425  \\
CNN     &         0.742  &         0.941  &         0.030  &         0.332  &                 0.388   \\
Bi-GRU  &         0.780  &         0.954  &         0.024  &         0.359  &                 0.420  \\ \midrule
CAML    &         0.820  & \textbf{0.966}* &         0.048  &         0.442  & \textbf{0.523}* \\
DR-CAML & \textbf{0.826} & \textbf{0.966}* & \textbf{0.049} & \textbf{0.457}* &                 0.515  \\
\bottomrule
\end{tabular}
\caption{Results on MIMIC-II full, 5031 labels.}
\label{tab:mimic2}
\end{table*}

\begin{table}
  \centering
\begin{tabular}{lll}
\toprule
& & Highly \\
  Method & Informative & informative \\
  \midrule
CAML & 46 & 22 \\
Code Descriptions & 48 & 20 \\
Logistic Regression & 41 & 18 \\
CNN & 36 & 13 \\
  \bottomrule
\end{tabular}
\caption{Qualitative evaluation results. The columns show the number of examples (out of 100) for which each method was selected as ``informative'' or ``highly informative''.}
\label{tab:qual_results}
\end{table}

\section{Related Work}

\paragraph{Attentional Convolution for NLP}

CNNs have been successfully applied to tasks such as sentiment classification~\cite{kim2014convolutional} and language modeling~\cite{dauphin2016language}. Our work combines convolution with attention~\cite{bahdanau2014neural,yang2016hierarchical} to select the most relevant parts of the discharge summary. Other recent work has combined convolution and attention~\cite[e.g.,][]{allamanis2016convolutional,yin2016abcnn,dos2016attentive,yin2017attentive}.
Our attention mechanism is most similar to those of \newcite{yang2016hierarchical} and \newcite{allamanis2016convolutional}, in that we use context vectors to compute attention over specific locations in the text. Our work differs in that we compute separate attention weights for each label in our label space, which is better tuned to our goal of selecting locations in a document which are most important for predicting specific labels.

\paragraph{Automatic ICD coding}
ICD coding is a long-standing task in the medical informatics community, which has been approached with machine learning and handcrafted methods~\cite{scheurwegs2015data}. Many recent approaches, like ours, use unstructured text data as the only source of information~\cite[e.g.,][]{kavuluru2015empirical,subotin2014system}, though some incorporates structured data as well~\cite[e.g.,][]{scheurwegs2017selecting,wang2016diagnosis}. 
Most previous methods have either evaluated only on a strict subset of the full ICD label space~\cite{wang2016diagnosis}, relied on datasets that focus on a subset of medical scenarios~\cite{zhang2017enhancing}, or evaluated on data that are not publicly available, making direct comparison difficult~\cite{subotin2016method}. A recent shared task for ICD-10 coding focused on coding of death certificates in English and French \cite{neveol2017clef}. This dataset also contains shorter documents than those we consider, with an average of 18 tokens per certificate in the French corpus. We use the open-access MIMIC datasets containing de-identified, general-purpose records of intensive care unit stays at a single hospital. 

\newcite{perotte2013diagnosis} use ``flat'' and ``hierarchical'' SVMs; the former treats each code as an individual prediction, while the latter trains on child codes only if the parent code is present, and predicts on child codes only if the parent code was positively predicted. \newcite{scheurwegs2017selecting} use a feature selection approach to ICD-9 and ICD-10 classification, incorporating structured and unstructured text information from EHRs. They evaluate over various medical specialties and on the MIMIC-III dataset. We compare directly to their results on the full label set of MIMIC-III.

Other recent approaches have employed neural network architectures. 
\newcite{baumel2017multi} apply recurrent networks with hierarchical sentence and word attention (the HA-GRU) to classify ICD9 diagnosis codes while providing insights into the model decision process. Similarly, \newcite{shi2017towards} applied character-aware LSTMs to generate sentence representations from specific subsections of discharge summaries, and apply attention to form a soft matching between the representations and the top 50 codes. \newcite{prakash2017condensed} use memory networks that draw from discharge summaries as well as Wikipedia, to predict top-50 and top-100 codes. Another recent neural architecture is the Grounded Recurrent Neural Network~\cite{vani2017grounded}, which employs a modified GRU with dimensions dedicated to predicting the presence of individual labels. We compare directly with published results from all of these papers, except \newcite{vani2017grounded}, who evaluate on only a 5000 code subset of ICD-9. Empirically, the CAML architecture proposed in this paper yields stronger results across all experimental conditions. We attribute these improvements to the attention mechanism, which focuses on the most critical features for each code, rather than applying a uniform pooling operation for all codes. We also observed that convolution-based models are at least as effective, and significantly more computationally efficient, than recurrent neural networks such as the Bi-GRU.

\paragraph{Explainable text classification}
A goal of this work is that the code predictions be explainable from features of the text. Prior work has also emphasized explainability. \newcite{lei2016rationalizing} model ``rationales'' through a latent variable, which tags each word as relevant to the document label. \newcite{li2016visualizing} compute the salience of individual words by the derivative of the label score with respect to the word embedding.
\newcite{ribeiro2016should} use submodular optimization to select a subset of features that closely approximate a specific classification decision (this work is also notable for extensive human evaluations). In comparison to these approaches, we employ a relatively simple attentional architecture; this simplicity is motivated by the challenge of scaling to multi-label classification with thousands of possible labels. Other prior work has emphasized the use of attention for highlighting salient features of the text~\cite[e.g.,][]{rush2015neural,rocktaschel2015reasoning}, although these papers did not perform human evaluations of the interpretability of the features selected by the attention mechanism.
\section{Conclusions and Future Work}
We present CAML, a convolutional neural network for multi-label document classification, which employs an attention mechanism to adaptively pool the convolution output for each label, learning to identify highly-predictive locations for each label. CAML yields strong improvements over previous metrics on several formulations of the ICD-9 code prediction task, while providing satisfactory explanations for its predictions. Although we focus on a clinical setting, CAML is extensible without modification to other multi-label document tagging tasks, including ICD-10 coding. We see a number of directions for future work. From the linguistic side, we plan to integrate the document structure of discharge summaries in MIMIC-III, and to better handle non-standard writing and other sources of out-of-vocabulary tokens. From the application perspective, we plan to build models that leverage hierarchy of ICD codes~\cite{choi2016doctor}, and to attempt the more difficult task of predicting diagnosis and treatment codes for \emph{future} visits from discharge summaries.  

\paragraph{Acknowledgments}
Helpful feedback was provided by the anonymous reviewers, and by the members of the Georgia Tech Computational Linguistics lab. The project was partially supported by project HDTRA1-15-1-0019 from the Defense Threat Reduction Agency, by the National Science Foundation under awards IIS-1418511 and CCF-1533768, by the National Institutes of Health under awards 1R01MD011682-01 and R56HL138415, by Children's Healthcare of Atlanta, and by UCB.

\bibliography{0-naaclhlt2018.bib}

\begin{thebibliography}{}
\expandafter\ifx\csname natexlab\endcsname\relax\def\natexlab#1{#1}\fi

\bibitem[{Allamanis et~al.(2016)Allamanis, Peng, and
  Sutton}]{allamanis2016convolutional}
Miltiadis Allamanis, Hao Peng, and Charles Sutton. 2016.
\newblock A convolutional attention network for extreme summarization of source
  code.
\newblock In {\em International Conference on Machine Learning\/}. pages
  2091--2100.

\bibitem[{Avati et~al.(2017)Avati, Jung, Harman, Downing, Ng, and
  Shah}]{avati2017improving}
Anand Avati, Kenneth Jung, Stephanie Harman, Lance Downing, Andrew Ng, and
  Nigam~H. Shah. 2017.
\newblock Improving palliative care with deep learning.
\newblock {\em arXiv preprint arXiv:1711.06402\/} .

\bibitem[{Bahdanau et~al.(2015)Bahdanau, Cho, and Bengio}]{bahdanau2014neural}
Dzmitry Bahdanau, Kyunghyun Cho, and Yoshua Bengio. 2015.
\newblock Neural machine translation by jointly learning to align and
  translate.
\newblock In {\em International Conference on Learning Representations\/}.

\bibitem[{Baumel et~al.(2018)Baumel, Nassour-Kassis, Elhadad, and
  Elhadad}]{baumel2017multi}
Tal Baumel, Jumana Nassour-Kassis, Michael Elhadad, and No{\'e}mie Elhadad.
  2018.
\newblock Multi-label classification of patient notes a case study on {ICD}
  code assignment.
\newblock In {\em AAAI Workshop on Health Intelligence\/}.

\bibitem[{Birman-Deych et~al.(2005)Birman-Deych, Waterman, Yan, Nilasena,
  Radford, and Gage}]{birman2005accuracy}
Elena Birman-Deych, Amy~D. Waterman, Yan Yan, David~S. Nilasena, Martha~J.
  Radford, and Brian~F Gage. 2005.
\newblock Accuracy of {ICD-9-CM} codes for identifying cardiovascular and
  stroke risk factors.
\newblock {\em Medical care\/} 43(5):480--485.

\bibitem[{Choi et~al.(2016)Choi, Bahadori, Schuetz, Stewart, and
  Sun}]{choi2016doctor}
Edward Choi, Mohammad~Taha Bahadori, Andy Schuetz, Walter~F Stewart, and Jimeng
  Sun. 2016.
\newblock Doctor {AI}: Predicting clinical events via recurrent neural
  networks.
\newblock In {\em Machine Learning for Healthcare Conference\/}. pages
  301--318.

\bibitem[{Dauphin et~al.(2017)Dauphin, Fan, Auli, and
  Grangier}]{dauphin2016language}
Yann~N. Dauphin, Angela Fan, Michael Auli, and David Grangier. 2017.
\newblock \href{http://proceedings.mlr.press/v70/dauphin17a.html}{Language
  modeling with gated convolutional networks}.
\newblock In {\em Proceedings of the 34th International Conference on Machine
  Learning\/}. pages 933--941.
\newblock \url{http://proceedings.mlr.press/v70/dauphin17a.html}.

\bibitem[{de~Lima et~al.(1998)de~Lima, Laender, and
  Ribeiro-Neto}]{de1998hierarchical}
Luciano~R.S. de~Lima, Alberto~H.F. Laender, and Berthier~A. Ribeiro-Neto. 1998.
\newblock A hierarchical approach to the automatic categorization of medical
  documents.
\newblock In {\em Proceedings of the seventh international conference on
  Information and knowledge management\/}. ACM, pages 132--139.

\bibitem[{Denny et~al.(2010)Denny, Ritchie, Basford, Pulley, Bastarache,
  Brown-Gentry, Wang, Masys, Roden, and Crawford}]{denny2010phewas}
Joshua~C. Denny, Marylyn~D. Ritchie, Melissa~A. Basford, Jill~M. Pulley, Lisa
  Bastarache, Kristin Brown-Gentry, Deede Wang, Dan~R. Masys, Dan~M. Roden, and
  Dana~C. Crawford. 2010.
\newblock Phe{WAS}: demonstrating the feasibility of a phenome-wide scan to
  discover gene--disease associations.
\newblock {\em Bioinformatics\/} 26(9):1205--1210.

\bibitem[{dos Santos et~al.(2016)dos Santos, Tan, Xiang, and
  Zhou}]{dos2016attentive}
C{\i}cero~Nogueira dos Santos, Ming Tan, Bing Xiang, and Bowen Zhou. 2016.
\newblock Attentive pooling networks.
\newblock {\em CoRR, abs/1602.03609\/} .

\bibitem[{Johnson et~al.(2016)Johnson, Pollard, Shen, Lehman, Feng, Ghassemi,
  Moody, Szolovits, Celi, and Mark}]{johnson2016mimic}
Alistair~E.W. Johnson, Tom~J. Pollard, Lu~Shen, Li-wei~H. Lehman, Mengling
  Feng, Mohammad Ghassemi, Benjamin Moody, Peter Szolovits, Leo~Anthony Celi,
  and Roger~G. Mark. 2016.
\newblock {MIMIC-III}, a freely accessible critical care database.
\newblock {\em Scientific data\/} 3.

\bibitem[{Kavuluru et~al.(2015)Kavuluru, Rios, and Lu}]{kavuluru2015empirical}
Ramakanth Kavuluru, Anthony Rios, and Yuan Lu. 2015.
\newblock An empirical evaluation of supervised learning approaches in
  assigning diagnosis codes to electronic medical records.
\newblock {\em Artificial intelligence in medicine\/} 65(2):155--166.

\bibitem[{Kim(2014)}]{kim2014convolutional}
Yoon Kim. 2014.
\newblock Convolutional neural networks for sentence classification.
\newblock In {\em Proceedings of the 2017 Conference on Empirical Methods in
  Natural Language Processing\/}. pages 1746--1751.

\bibitem[{Kingma and Ba(2015)}]{kingma2014adam}
Diederik Kingma and Jimmy Ba. 2015.
\newblock Adam: A method for stochastic optimization.
\newblock In {\em International Conference on Learning Representations\/}.

\bibitem[{Lei et~al.(2016)Lei, Barzilay, and Jaakkola}]{lei2016rationalizing}
Tao Lei, Regina Barzilay, and Tommi Jaakkola. 2016.
\newblock \href{https://aclweb.org/anthology/D16-1011}{Rationalizing neural
  predictions}.
\newblock In {\em Proceedings of the 2016 Conference on Empirical Methods in
  Natural Language Processing\/}. Association for Computational Linguistics,
  Austin, Texas, pages 107--117.
\newblock \url{https://aclweb.org/anthology/D16-1011}.

\bibitem[{Li et~al.(2016)Li, Chen, Hovy, and Jurafsky}]{li2016visualizing}
Jiwei Li, Xinlei Chen, Eduard Hovy, and Dan Jurafsky. 2016.
\newblock \href{http://www.aclweb.org/anthology/N16-1082}{Visualizing and
  understanding neural models in {NLP}}.
\newblock In {\em Proceedings of the 2016 Conference of the North American
  Chapter of the Association for Computational Linguistics: Human Language
  Technologies\/}. Association for Computational Linguistics, San Diego,
  California, pages 681--691.
\newblock \url{http://www.aclweb.org/anthology/N16-1082}.

\bibitem[{McCallum(1999)}]{mccallum1999multi}
Andrew McCallum. 1999.
\newblock Multi-label text classification with a mixture model trained by {EM}.
\newblock In {\em AAAI workshop on Text Learning\/}. pages 1--7.

\bibitem[{Mikolov et~al.(2013)Mikolov, Sutskever, Chen, Corrado, and
  Dean}]{mikolov2013efficient}
Tomas Mikolov, Ilya Sutskever, Kai Chen, Greg~S Corrado, and Jeff Dean. 2013.
\newblock Distributed representations of words and phrases and their
  compositionality.
\newblock In {\em Advances in neural information processing systems\/}. pages
  3111--3119.

\bibitem[{N{\'e}v{\'e}ol et~al.(2017)N{\'e}v{\'e}ol, Anderson, Cohen, Grouin,
  Lavergne, Rey, Robert, Rondet, and Zweigenbaum}]{neveol2017clef}
Aur{\'e}lie N{\'e}v{\'e}ol, Robert~N Anderson, K~Bretonnel Cohen, Cyril Grouin,
  Thomas Lavergne, Gr{\'e}goire Rey, Aude Robert, Claire Rondet, and Pierre
  Zweigenbaum. 2017.
\newblock {CLEF} ehealth 2017 multilingual information extraction task
  overview: {ICD10} coding of death certificates in english and french.
\newblock In {\em CLEF 2017 Evaluation Labs and Workshop: Online Working Notes,
  CEUR-WS\/}. page~17.

\bibitem[{Perotte et~al.(2013)Perotte, Pivovarov, Natarajan, Weiskopf, Wood,
  and Elhadad}]{perotte2013diagnosis}
Adler Perotte, Rimma Pivovarov, Karthik Natarajan, Nicole Weiskopf, Frank Wood,
  and No{\'e}mie Elhadad. 2013.
\newblock Diagnosis code assignment: models and evaluation metrics.
\newblock {\em Journal of the American Medical Informatics Association\/}
  21(2):231--237.

\bibitem[{Prakash et~al.(2017)Prakash, Zhao, Hasan, Datla, Lee, Qadir, Liu, and
  Farri}]{prakash2017condensed}
Aaditya Prakash, Siyuan Zhao, Sadid~A Hasan, Vivek~V Datla, Kathy Lee, Ashequl
  Qadir, Joey Liu, and Oladimeji Farri. 2017.
\newblock Condensed memory networks for clinical diagnostic inferencing.
\newblock In {\em Proceedings of the Thirty-First AAAI Conference on Artificial
  Intelligence\/}. pages 3274--3280.

\bibitem[{Ranganath et~al.(2015)Ranganath, Perotte, Elhadad, and
  Blei}]{ranganath2015survival}
Rajesh Ranganath, Adler~J. Perotte, No{\'e}mie Elhadad, and David~M. Blei.
  2015.
\newblock The survival filter: Joint survival analysis with a latent time
  series.
\newblock In {\em UAI\/}. pages 742--751.

\bibitem[{Ribeiro et~al.(2016)Ribeiro, Singh, and Guestrin}]{ribeiro2016should}
Marco~Tulio Ribeiro, Sameer Singh, and Carlos Guestrin. 2016.
\newblock Why should i trust you?: Explaining the predictions of any
  classifier.
\newblock In {\em Proceedings of the 22nd ACM SIGKDD International Conference
  on Knowledge Discovery and Data Mining\/}. ACM, pages 1135--1144.

\bibitem[{Rockt{\"a}schel et~al.(2016)Rockt{\"a}schel, Grefenstette, Hermann,
  Ko{\v{c}}isk{\`y}, and Blunsom}]{rocktaschel2015reasoning}
Tim Rockt{\"a}schel, Edward Grefenstette, Karl~Moritz Hermann, Tom{\'a}{\v{s}}
  Ko{\v{c}}isk{\`y}, and Phil Blunsom. 2016.
\newblock Reasoning about entailment with neural attention.
\newblock In {\em International Conference on Learning Representations\/}.

\bibitem[{Rush et~al.(2015)Rush, Chopra, and Weston}]{rush2015neural}
Alexander~M. Rush, Sumit Chopra, and Jason Weston. 2015.
\newblock \href{http://aclweb.org/anthology/D15-1044}{A neural attention model
  for abstractive sentence summarization}.
\newblock In {\em Proceedings of the 2015 Conference on Empirical Methods in
  Natural Language Processing\/}. Association for Computational Linguistics,
  Lisbon, Portugal, pages 379--389.
\newblock \url{http://aclweb.org/anthology/D15-1044}.

\bibitem[{Scheurwegs et~al.(2017)Scheurwegs, Cule, Luyckx, Luyten, and
  Daelemans}]{scheurwegs2017selecting}
Elyne Scheurwegs, Boris Cule, Kim Luyckx, L{\'e}on Luyten, and Walter
  Daelemans. 2017.
\newblock Selecting relevant features from the electronic health record for
  clinical code prediction.
\newblock {\em Journal of Biomedical Informatics\/} 74:92--103.

\bibitem[{Scheurwegs et~al.(2015)Scheurwegs, Luyckx, Luyten, Daelemans, and
  Van~den Bulcke}]{scheurwegs2015data}
Elyne Scheurwegs, Kim Luyckx, L{\'e}on Luyten, Walter Daelemans, and Tim
  Van~den Bulcke. 2015.
\newblock Data integration of structured and unstructured sources for assigning
  clinical codes to patient stays.
\newblock {\em Journal of the American Medical Informatics Association\/}
  23(e1):e11--e19.

\bibitem[{Shi et~al.(2017)Shi, Xie, Hu, Zhang, and Xing}]{shi2017towards}
Haoran Shi, Pengtao Xie, Zhiting Hu, Ming Zhang, and Eric~P Xing. 2017.
\newblock Towards automated {ICD} coding using deep learning.
\newblock {\em arXiv preprint arXiv:1711.04075\/} .

\bibitem[{Snoek et~al.(2012)Snoek, Larochelle, and Adams}]{NIPS2012_4522}
Jasper Snoek, Hugo Larochelle, and Ryan~P Adams. 2012.
\newblock Practical bayesian optimization of machine learning algorithms.
\newblock In F.~Pereira, C.~J.~C. Burges, L.~Bottou, and K.~Q. Weinberger,
  editors, {\em Advances in Neural Information Processing Systems 25\/}, Curran
  Associates, Inc., pages 2951--2959.

\bibitem[{Subotin and Davis(2014)}]{subotin2014system}
Michael Subotin and Anthony~R. Davis. 2014.
\newblock A system for predicting {ICD-10-PCS} codes from electronic health
  records.
\newblock In {\em Proceedings of the 2014 Workshop on Biomedical Natural
  Language Processing\/}.

\bibitem[{Subotin and Davis(2016)}]{subotin2016method}
Michael Subotin and Anthony~R. Davis. 2016.
\newblock A method for modeling co-occurrence propensity of clinical codes with
  application to {ICD-10-PCS} auto-coding.
\newblock {\em Journal of the American Medical Informatics Association\/}
  23(5):866--871.

\bibitem[{Swersky et~al.(2013)Swersky, Snoek, and Adams}]{NIPS2013_5086}
Kevin Swersky, Jasper Snoek, and Ryan~P Adams. 2013.
\newblock Multi-task bayesian optimization.
\newblock In C.~J.~C. Burges, L.~Bottou, M.~Welling, Z.~Ghahramani, and K.~Q.
  Weinberger, editors, {\em Advances in Neural Information Processing Systems
  26\/}, Curran Associates, Inc., pages 2004--2012.

\bibitem[{Vani et~al.(2017)Vani, Jernite, and Sontag}]{vani2017grounded}
Ankit Vani, Yacine Jernite, and David Sontag. 2017.
\newblock Grounded recurrent neural networks.
\newblock {\em arXiv preprint arXiv:1705.08557\/} .

\bibitem[{Wang et~al.(2016)Wang, Chang, Li, Long, Yao, and
  Sheng}]{wang2016diagnosis}
Sen Wang, Xiaojun Chang, Xue Li, Guodong Long, Lina Yao, and Quan~Z Sheng.
  2016.
\newblock Diagnosis code assignment using sparsity-based disease correlation
  embedding.
\newblock {\em IEEE Transactions on Knowledge and Data Engineering\/}
  28(12):3191--3202.

\bibitem[{{World Health Organization}(2016)}]{ICD10}
{World Health Organization}. 2016.
\newblock
  \href{http://apps.who.int/classifications/icd10/browse/2016/en}{International
  statistical classification of diseases and related health problems 10th
  revision}.
\newblock \url{http://apps.who.int/classifications/icd10/browse/2016/en}.

\bibitem[{Yang et~al.(2016)Yang, Yang, Dyer, He, Smola, and
  Hovy}]{yang2016hierarchical}
Zichao Yang, Diyi Yang, Chris Dyer, Xiaodong He, Alexander~J. Smola, and
  Eduard~H. Hovy. 2016.
\newblock Hierarchical attention networks for document classification.
\newblock In {\em Proceedings of the 2016 Conference of the North American
  Chapter of the Association for Computational Linguistics: Human Language
  Technologies\/}. Association for Computational Linguistics, San Diego,
  California, pages 1480--1489.

\bibitem[{Yin and Sch{\"u}tze(2017)}]{yin2017attentive}
Wenpeng Yin and Hinrich Sch{\"u}tze. 2017.
\newblock Attentive convolution.
\newblock {\em arXiv preprint arXiv:1710.00519\/} .

\bibitem[{Yin et~al.(2016)Yin, Sch{\"u}tze, Xiang, and Zhou}]{yin2016abcnn}
Wenpeng Yin, Hinrich Sch{\"u}tze, Bing Xiang, and Bowen Zhou. 2016.
\newblock {ABCNN}: Attention-based convolutional neural network for modeling
  sentence pairs.
\newblock {\em Transactions of the Association for Computational Linguistics\/}
  4:259--272.

\bibitem[{Zhang et~al.(2017)Zhang, He, Zhao, and Li}]{zhang2017enhancing}
Danchen Zhang, Daqing He, Sanqiang Zhao, and Lei Li. 2017.
\newblock Enhancing automatic {ICD-9-CM} code assignment for medical texts with
  pubmed.
\newblock {\em BioNLP 2017\/} pages 263--271.

\end{thebibliography}
\bibliographystyle{acl_natbib}

\end{document}